\documentclass[10pt,twocolumn,letterpaper]{article}

\usepackage{cvpr}              
\definecolor{cvprblue}{rgb}{0.21,0.49,0.74}
\usepackage[pagebackref,breaklinks,colorlinks,allcolors=cvprblue]{hyperref}

\usepackage[T1]{fontenc}         
\usepackage[utf8]{inputenc}      
\usepackage{xcolor}              
\usepackage{amsmath, amssymb}    
\usepackage{graphicx}            
\usepackage{booktabs}            
\usepackage{microtype}           
\usepackage{url} 
\usepackage{pifont} 
\usepackage{hyperref}            
\usepackage{booktabs}    
\usepackage{colortbl}    
\usepackage{xcolor}      
\usepackage{adjustbox}   
\usepackage{multirow}
\usepackage{pifont}
\usepackage{comment}
\usepackage{hyperref}
\usepackage{bbm} 
\usepackage[most]{tcolorbox}

\newcommand{\ind}{\mathbbm{1}} 
\usepackage{xcolor} 
\definecolor{ourteal}{RGB}{0,128,128} 

\def\confName{CVPR}
\def\confYear{2026}

\title{
\raisebox{-0.35\height}{\includegraphics[height=2em]{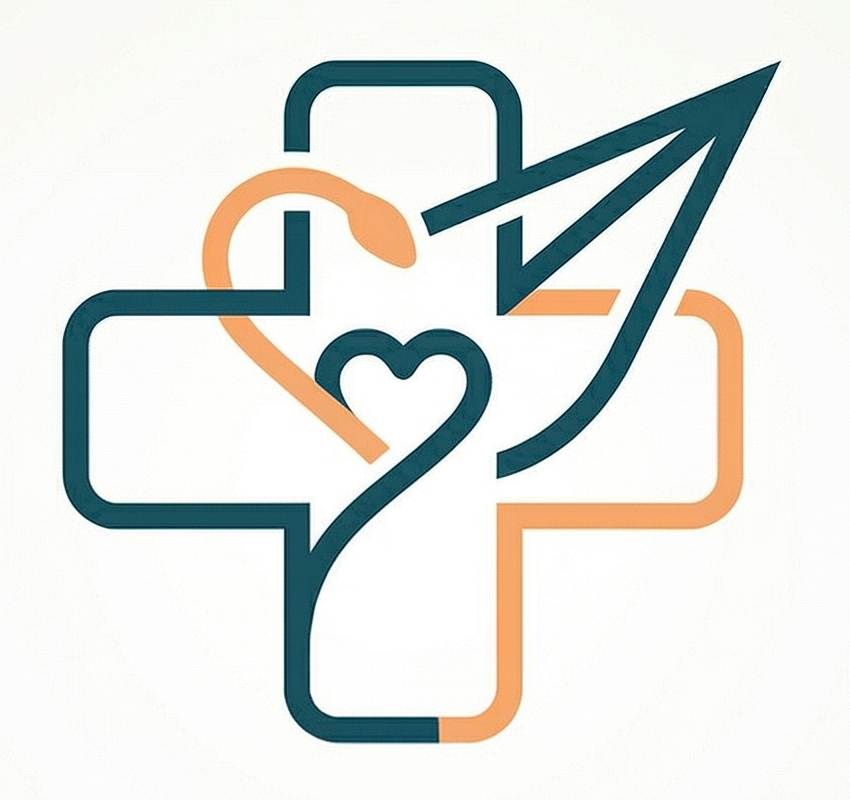}}
\hspace{-0.5em}
\textit{CarePilot:} A Multi-Agent Framework for Long-Horizon Computer Task Automation in Healthcare
}

\author{%
  Akash Ghosh$^{1}$\thanks{Work done as Visiting Researcher at MBZUAI} \quad
  Tajamul Ashraf$^{2}$ \quad
  Rishu Kumar Singh$^{1}$ \quad
  Numan Saeed$^{2}$ \\
  Sriparna Saha$^{1}$ \quad
  Xiuying Chen$^{2}$ \thanks{Corresponding Author} \quad
  Salman Khan$^{2}$ \\[0.4em]
  {\small
    $^{1}$Indian Institute of Technology Patna \quad
    $^{2}$Mohamed bin Zayed University of AI (MBZUAI)
  } \\[0.2em]
}

\begin{document}

\maketitle
\begin{abstract}
Multimodal agentic pipelines are transforming human–computer interaction by enabling efficient and accessible automation of complex, real-world tasks. However, recent efforts have focused on short-horizon or general-purpose applications (e.g., mobile or desktop interfaces), leaving long-horizon automation for domain-specific systems, particularly in healthcare, largely unexplored.
To address this, we introduce CareFlow, a high-quality human-annotated benchmark comprising complex, long-horizon software workflows across medical annotation tools, DICOM viewers, EHR systems, and laboratory information systems. On this benchmark, existing vision–language models (VLMs) perform poorly, struggling with long-horizon reasoning and multi-step interactions in medical contexts.
To overcome this, we propose \textbf{\textit{CarePilot}}, a multi-agent framework based on the actor–critic paradigm. The Actor integrates tool grounding with dual-memory mechanisms, long-term and short-term experience, to predict the next semantic action from the visual interface and system state. The Critic evaluates each action, updates memory based on observed effects, and either executes or provides corrective feedback to refine the workflow. Through iterative agentic simulation, the Actor learns to perform more robust and reasoning-aware predictions during inference. Our experiments show that \textbf{\textit{CarePilot}} achieves state-of-the-art performance, outperforming strong closed-source and open-source multimodal baselines by approximately \textbf{15.26\%} and \textbf{3.38\%}, on our benchmark and out-of-distribution dataset, respectively. The code and dataset for this project are available at: \href{https://akashghosh.github.io/Care-Pilot/}{Carepilot}.
\end{abstract} 
\vspace{-0.5em}
\section{Introduction}
\label{sec:intro}

\begin{figure}
    \centering
    \includegraphics[width=\linewidth]{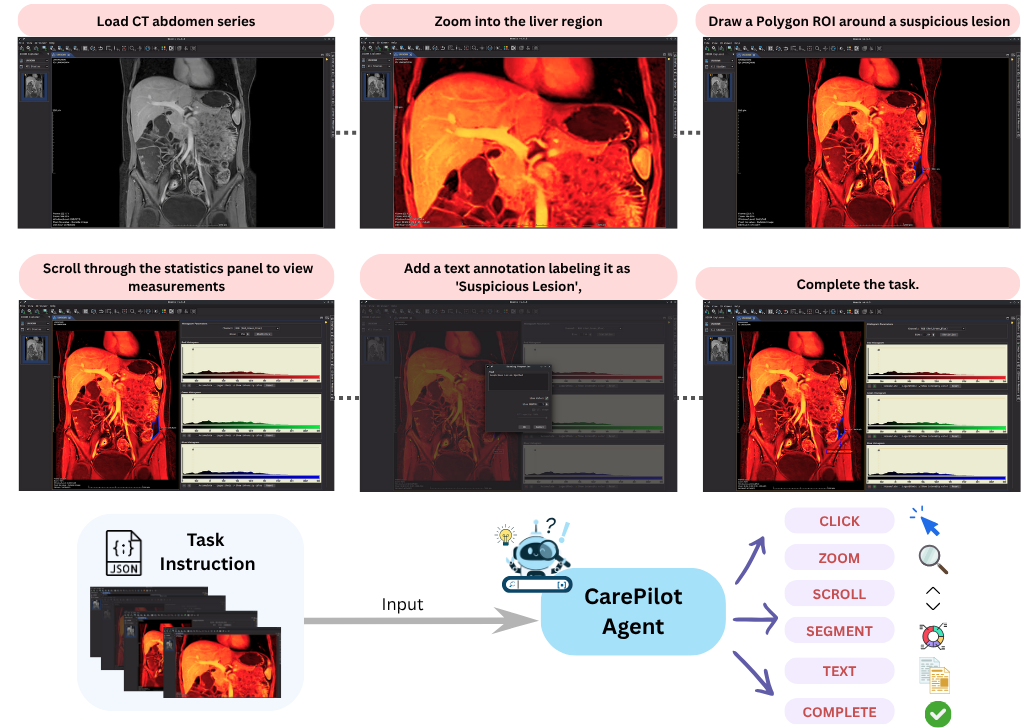}
    \caption{\textbf{CareFlow} is a large-scale benchmark for evaluating multimodal agents on a range of real healthcare software. It enables execution-based evaluation and interactive reasoning across DICOM viewers, image annotation tools, EMR/EHR, and LIS platforms. Each task pairs a natural-language goal with GUI screenshots representing authentic clinical workflows.}

    \label{fig:placeholder}
\end{figure}
In today’s digital world, software systems powered by large language models (LLMs) and vision language models (VLMs) form the backbone of modern activity, shaping how humans learn, collaborate, analyze data, and create content across domains, interfaces, and tools~\cite{ashraf2025agentxevaluatingdeepmultimodal, mialon2023augmented, liang2024taskmatrix,ghosh2024exploring,ghosh2024clipsyntel,ghosh2025survey,ghosh2024medsumm,ghosh2024sights,ghosh2024healthalignsumm,ghosh2025infogen,acharya2025m3retrieve,maji2025sanskriti,maji2025drishtikon,dmaster, Ashraf_2024_WACV}. Recent advances in large multimodal models (LLMs) have enabled autonomous software agents that can follow high-level natural language instructions to operate real applications, making complex computer use more accessible and efficient~\cite{zhang2024large,zheng2024gpt}. However, building agents that can reliably execute long-horizon workflows, spanning dozens of interdependent steps under partial observability, remains a major challenge. A key obstacle is the lack of realistic and interactive benchmarks that capture the heterogeneity of operating systems, interfaces, and domain-specific software environments~\cite{xie2024osworld, schmidgall2024agentclinic}. Moreover, such long-horizon multimodal agents require robust grounding and memory mechanisms to make informed decisions at each step, a difficult problem that demands effective use of tool representations, contextual reasoning, and long-term memory integration \cite{wang2023voyager,shinn2023reflexion,wang2024jarvis,onyame2026cure}.

Healthcare software ecosystems are inherently broad and workflow-centric, spanning DICOM servers/viewers, image-computing and annotation tools, EMR/EHR systems, and laboratory information systems (LIS)~\cite{annotationTools2025,monai2025multimodal}. Day-to-day clinical use often requires chaining 10--15 dependent actions, for example, opening a study, configuring views, annotating or measuring, exporting artifacts, and updating records, while adhering to data integrity, audit trails, and strict privacy policies. These platforms are highly heterogeneous and policy-constrained, and they evolve frequently: user-interface updates, custom deployments, and institution-specific configurations make agents that overfit to surface layouts brittle. This combination of heterogeneity, long-horizon dependencies, and strict compliance requirements makes healthcare a natural yet uniquely challenging testbed for long-horizon GUI agents~\cite{awesomeGUIAgent2025}.

Despite recent progress on long-horizon multimodal agents in Android, desktop, and web environments~\cite{toyama2021androidenv, zhang2025appagent, xie2024osworld, medspot}, there remains no standardized public benchmark for healthcare or clinical settings that reflects how users interact with multiple medical softwares. This absence of domain-grounded evaluation makes it difficult to assess how current agents generalize to healthcare-specific tasks typically performed by trained medical professionals
~\cite{schmidgall2024agentclinic}. Addressing this gap is essential for developing robust, trustworthy multimodal agents capable of operating safely and efficiently in clinical software ecosystems.

With this motivation, we introduce {\bf CareFlow}, a healthcare-specific long-horizon benchmark that evaluates complex workflows requiring domain knowledge of specialized software. {\bf CareFlow} contains tasks spanning 8–24 consecutive decisions, executed over sequences of GUI screenshots from real medical softwares. At each timestep $t$, the agent receives the current screenshot, the task instruction, and a condensed history of prior states/actions, and must predict the next semantic action to advance the workflow.

A key challenge in building such benchmarks lies in constructing long-horizon queries that are both high-quality and faithfully reflect real-world software usage. To ensure realism, we collaborated closely with domain experts to draft seed instructions covering the core operations they routinely perform. For each instruction, we recorded detailed step-by-step workflows required to complete the corresponding task. We then filtered and refined these trajectories to retain high-frequency, high-value procedures that are critical in everyday clinical practice. For example, in medical image annotation, we focused on 3D~Slicer~\cite{fedorov20123d}, one of the most widely adopted open-source tools for volumetric analysis, and curated representative workflows for annotation, segmentation, and measurement tasks.

To enable multimodal LLMs to tackle complex, domain-specific, long-horizon workflows in healthcare software ecosystems, we propose \textbf{\textit{CarePilot}}, a memory- and tool-augmented multi-agent framework inspired by the actor–critic paradigm~\cite{iqbal2019actorattentioncritic}. At time step \(t\), the \textbf{Actor} (a multimodal LM) receives the current screenshot and instruction, invokes lightweight tool modules (e.g., zoom/crop, OCR, UI/object detection) to obtain grounding signals, and predicts the next semantic action. A dual-memory design underpins the system: the \emph{long-term memory} compacts the history up to \(t\!-\!1\) (key states, actions, outcomes), and the \emph{short-term memory} records the most recent decision and feedback at time \(t\). The \textbf{Critic} evaluates the Actor’s proposal, updates both memories with observed effects, and issues corrective feedback, during training comparing the Actor’s action to reference traces and during evaluation relying on execution outcomes or verifier feedback. If accepted, the action advances the workflow; if revised, the Actor re-plans. At time \(t\!+\!1\), the Actor conditions on the refreshed memory and grounding signals to produce a more informed action. 

Our work makes the following key contributions:

\begin{itemize}
    \item \textbf{Problem Formulation.} We define a new task of \emph{long-horizon computer automation for healthcare software}: given a natural-language goal and a sequence of screenshots, an agent must predict step-by-step actions to complete real clinical workflows.
    
    \item \textbf{Benchmark.} We present \textbf{CareFlow}, an expert-annotated benchmark of long-horizon healthcare workflows comprising 8--24 steps for each task encompassing four major clinical systems. Each task is labeled with interface invariant semantic actions and verified using artifact/state based checks.
    
    \item \textbf{Framework.} We propose \textbf{\textit{CarePilot}}, a multi-agent framework built on the actor–critic paradigm that integrates tool grounding with dual memories (long and short-term) for robust next-action prediction.
    
    \item \textbf{Evaluation.} Extensive experiments across all \textbf{{CareFlow}} domains show that \textbf{\textit{CarePilot}} achieves state-of-the-art results, improving  task accuracy upto \textbf{15.26\%} over strong open- and closed-source baselines. 

\end{itemize}

\section{Related Work}
\begin{figure*}[t]
    \centering
    \includegraphics[width=\linewidth]{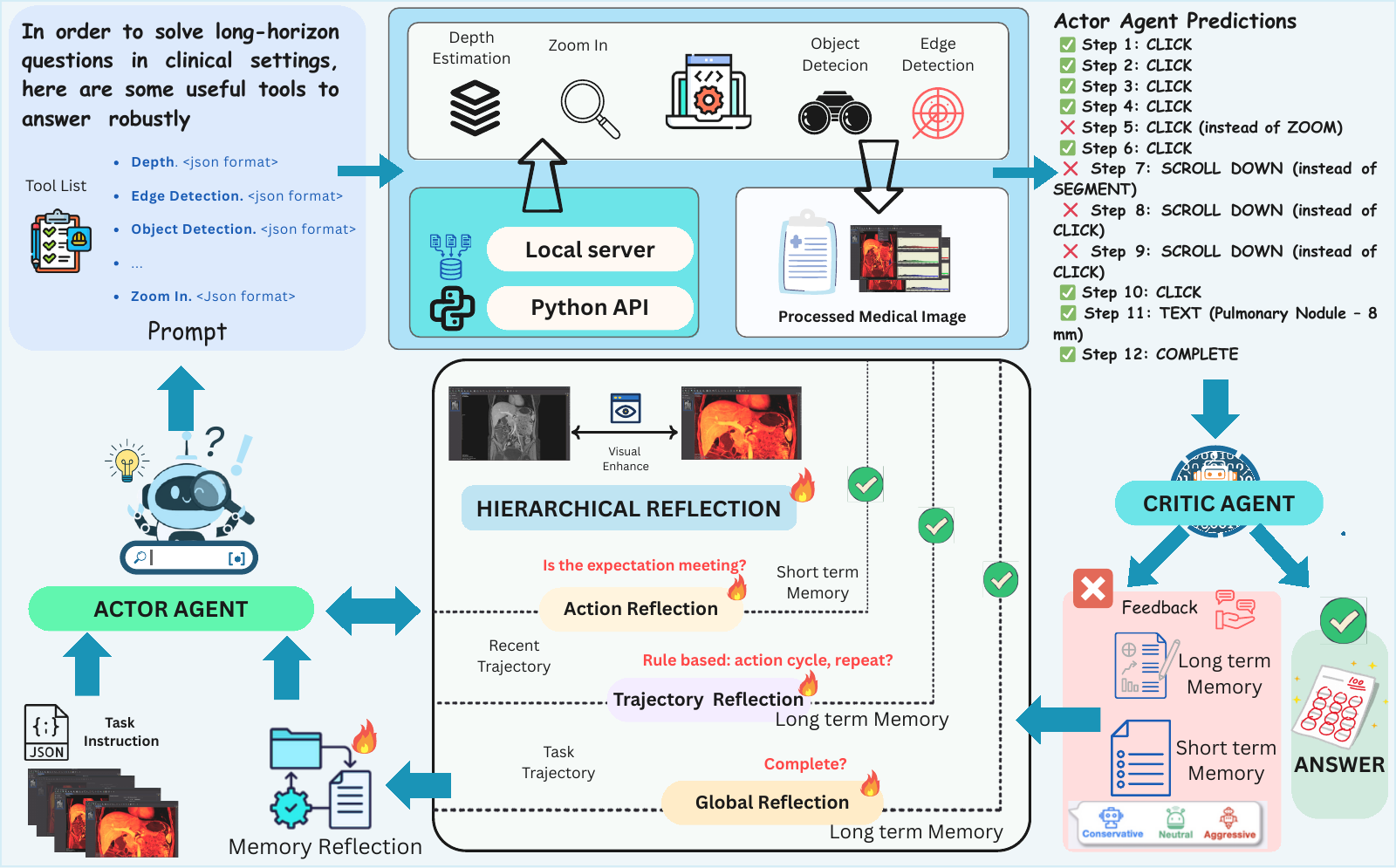}
    \caption{\textbf{Overview of the \textit{CarePilot}} framework. 
An Actor–Critic multi-agent architecture governs hierarchical decision-making for long-horizon healthcare workflows. 
At each step, the \textbf{Actor} observes the current interface and instruction, integrates tool-grounding signals, and its past experience that is stored in short- and long-term memories, and predicts the next semantic action.  The \textbf{Critic} evaluates outcomes, provides corrective feedback, and updates both short-term and long-term memory buffers to guide subsequent decisions.}
    \label{fig:main_flow}
\end{figure*}
\subsection{Autonomous Multimodal Agents}
Recent advances in multimodal agents have enabled models to perceive, reason, and act within digital environments by grounding visual and textual inputs into executable actions. Systems such as Mind2Web~\cite{deng2023mind2web}, SeeAct~\cite{zheng2024gpt}, and UI-TARS~\cite{qin2025ui} leverage screenshot-based reasoning and instructions to automate interactions across web and desktop applications. Large-scale benchmarks including WebArena~\cite{zhou2023webarena} and AppWorld~\cite{trivedi2024appworld}, further extend these capabilities to diverse real-world contexts. However, these efforts primarily target short-horizon, general-purpose tasks where domain-specific reasoning remains limited.
To improve temporal coherence and planning, several works have explored memory-augmented and actor–critic-based agents. Voyager~\cite{wang2023voyager}, Reflexion~\cite{shinn2023reflexion}, and Jarvis 1~\cite{wang2024jarvis} demonstrate the importance of episodic memory, self-reflection, and long-term credit assignment for persistent task execution.
However, the medical domain still lacks agentic systems capable of operating in real-world clinical environments to assist in downstream tasks such as diagnosis, workflow optimization, and decision support. Existing approaches primarily focus on general or robotic settings, with limited emphasis on clinical reasoning and safety-critical adaptability.
Building on these insights, our proposed {\bf \textit{CarePilot}} introduces a dual-memory actor–critic framework that couples long-horizon experience replay with short-term contextual grounding. This design enables robust, reasoning-aware action prediction and adaptive correction across complex, multi-step healthcare workflows.

\subsection{Healthcare Software Automation}
Automation in healthcare software has largely relied on rule-based or heuristic-driven systems for electronic medical record (EMR/EHR) management, DICOM image visualization, and laboratory information processing~\cite{rajkomar2018scalable, miotto2018deep}. While such methods improve efficiency, they lack generalization across heterogeneous clinical interfaces and cannot reason over multi-stage tasks. Recent multimodal medical AI systems have emphasized perception, such as diagnostic imaging~\cite{azad2023foundational} and report generation~\cite{li2025multimodal}, but have not addressed interactive software control.
{\bf{CareFlow}} bridges this gap by introducing a fully human-annotated benchmark of long-horizon healthcare software interactions, covering EMR systems, annotation tools, and hospital management applications. Together with {\bf{\textit{CarePilot}}}, this constitutes the first end-to-end multimodal agentic framework that perceives, reasons, and acts within complex healthcare software ecosystems, paving the way toward safe, interpretable, and generalizable automation in clinical environments.




\section{CareFlow}
To systematically evaluate multimodal LLMs on long-horizon healthcare tasks, we introduce \textbf{CareFlow}, a high-quality benchmark of real-world software workflows. This section details the benchmark’s composition, statistics, and characteristics, and describes the complete annotation pipeline used to construct \textbf{CareFlow} (Figure~\ref{fig:main_flow}). 

\begin{figure}[t]
    \centering
    \includegraphics[width=\linewidth]{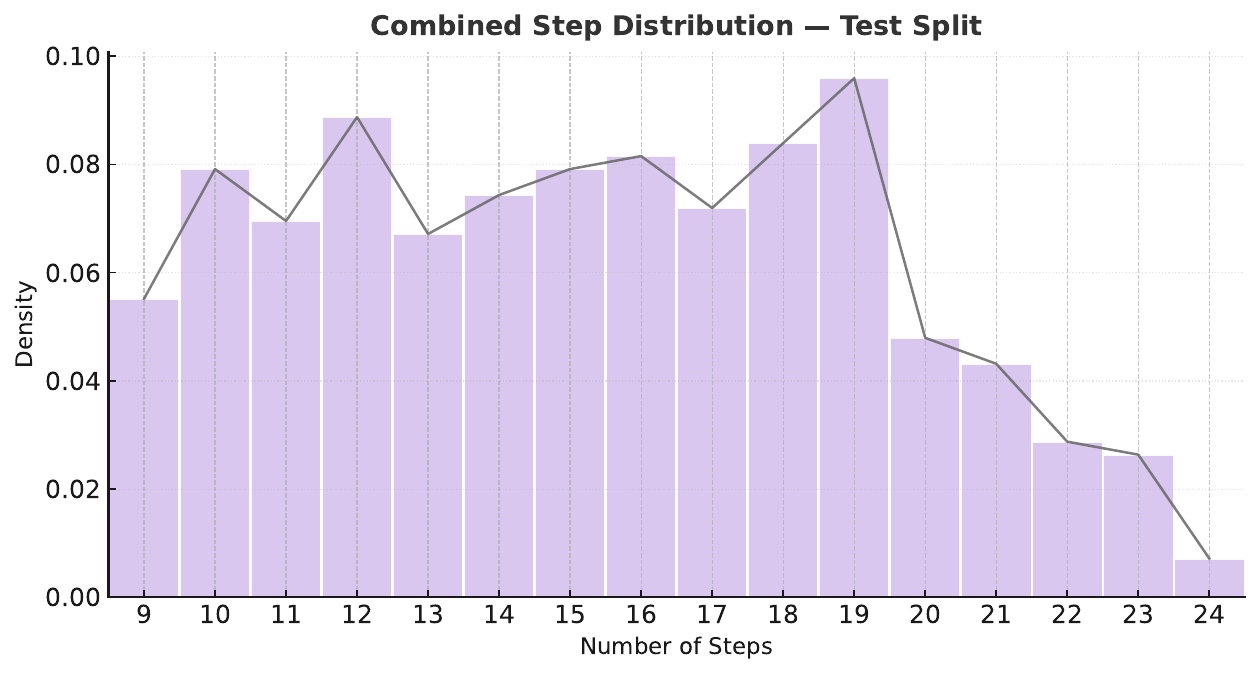}
    \caption{Distribution of task lengths (number of steps) in the test split of \textbf{\textit{CareFlow}}.}
    \label{fig:step_distribution}
\end{figure}

\subsection{Dataset Pipeline}
The \textbf{CareFlow} dataset is constructed through a carefully designed four-stage annotation pipeline to ensure diversity, realism, and reproducibility of healthcare workflows.

\textbf{(i) Crafting Seed Tasks.} We collaborated with domain experts to map each software system’s real-world usage patterns, functional scope, and operational constraints. Through brainstorming sessions, we identified the core activities performed by practitioners and distilled a seed inventory of executable, end-to-end tasks representative of authentic clinical workflows.

\textbf{(ii) Expanding Diversity and Scale.} To broaden coverage and increase sample count, we systematically generated diverse variants of each seed instruction. These variations included controlled substitutions (e.g., replacing ``MRI report'' with ``X-ray report''), parameter adjustments (filenames, thresholds), and procedural edits such as adding or omitting optional zoom or configuration steps, while preserving intent and executability.

\textbf{(iii) Stepwise Annotation of GUI States.} Each generated task was decomposed into a clear sequence of atomic steps by trained annotators. For every step, annotators captured the corresponding screenshot and labeled the precise next semantic action required to progress within the interface. This produced fully grounded, screenshot–action pairs for long-horizon reasoning.

\textbf{(iv) Quality Assurance and Filtering.} We retained only those trajectories that met three strict criteria: (a) chronological consistency of screenshots, (b) task completeness with optimal or near-optimal step sequences, and (c) clear, unambiguous natural-language instructions. Any instance failing one or more of these checks was discarded.\par
The entire annotation process was supervised by two domain experts who routinely use these healthcare software systems, while two trained interns populated the images and task formulations under their guidance. The test set was independently validated by the experts, and inter-annotator agreement, measured using Cohen's kappa ($\kappa$), was $0.78$.

\begin{figure}[t]
    \centering
    \includegraphics[width=0.58\linewidth]{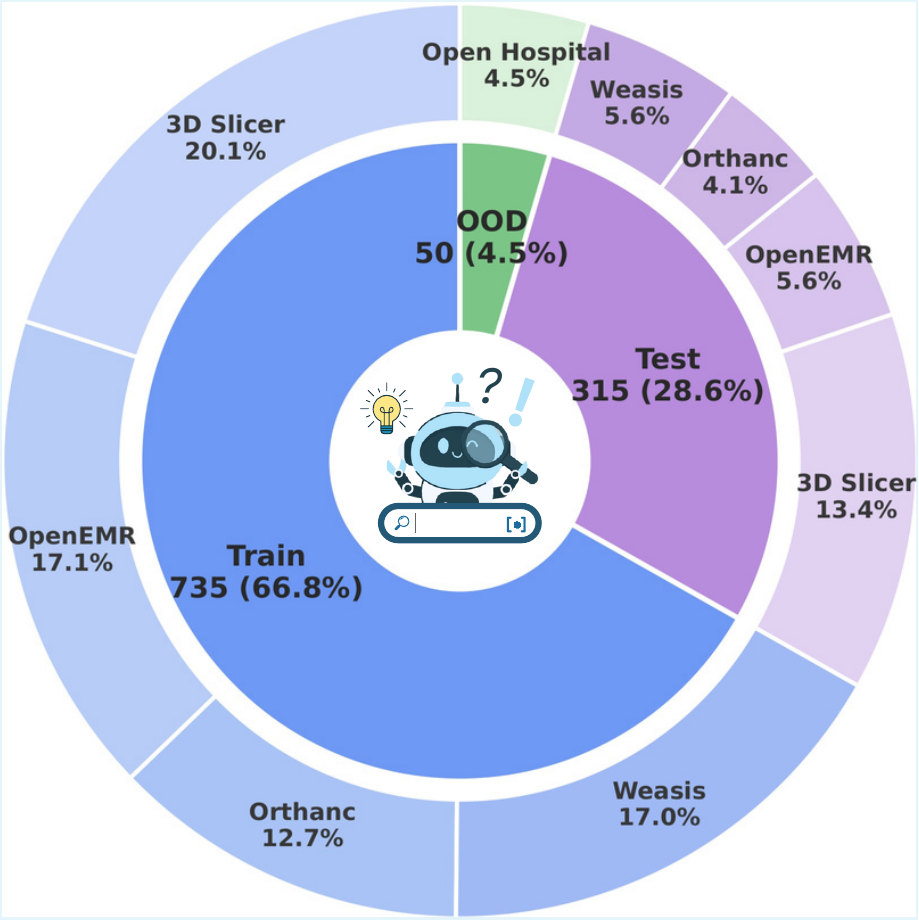}
    \caption{Category distribution of tasks in \textbf{\textit{CareFlow}} across four major healthcare software domains.}
    \label{fig:careflow_distribution}
\end{figure}

\subsection{Dataset Characteristics}
\label{sec:dataset_characteristics}

\textbf{CareFlow} spans four major categories of healthcare software: 
(i) DICOM viewing and infrastructure (\href{https://www.orthanc-server.com/}{Orthanc}, \href{https://weasis.org/}{Weasis}), (ii) medical image computing and annotation (\href{https://www.slicer.org}{3D-Slicer}), (iii) hospital information and EMR systems (\href{https://www.open-emr.org}{OpenEMR}), and 
(iv) laboratory information systems (\href{https://www.open-hospital.org}{OpenHospital}). 
The benchmark contains 1{,}100 tasks collected across these platforms, each defined by a complex natural-language instruction and a trajectory of 8--24 consecutive GUI screenshots. Each screenshot corresponds to the application state at time step~$t$ within a multi-step workflow. For every state~$t$, we provide an \emph{interface-invariant} next-action label in text, indicating the operation required at~$t{+}1$ to advance toward task completion. 
The action space of \textbf{\textit{CareFlow}} includes six core operations, \texttt{CLICK}, \texttt{SCROLL}, \texttt{ZOOM}, \texttt{TEXT}, \texttt{SEGMENT} and \texttt{COMPLETE} covering the primitive interactions needed for complex healthcare software workflows (see Table~\ref{tab:health-actions}). 
Figure~\ref{fig:careflow_distribution} illustrates the data composition across the five software categories, while Figure~\ref{fig:step_distribution} shows the distribution of task lengths.  This design aligns with recent multimodal GUI benchmarks like GUIOdyssey \cite{lu2025guiodyssey} while addressing the unique challenges of clinical systems.


\definecolor{headblue}{RGB}{232,240,255}
\definecolor{rowgray}{gray}{0.98}

\begin{table}[t]
\centering
\small
\setlength{\tabcolsep}{6pt}
\renewcommand{\arraystretch}{0.6}

\caption{Core mouse and keyboard actions in \textbf{\textit{CarePilot}}. 
Arguments in parentheses denote examples; $(x,y)$ are pixel coordinates, $n$ is a signed step count, and $s$ is a text string.}
\label{tab:health-actions}

\rowcolors{2}{rowgray}{white}
\resizebox{\columnwidth}{!}{%
\begin{tabular}{>{\ttfamily}l p{0.74\linewidth}}
\rowcolor{headblue}
\toprule
\textbf{Function} & \textbf{Description} \\
\toprule
CLICK & Move the cursor (if needed) and click at the specified item in the screen \\
SCROLL & Scroll the active view vertically or horizontally by $n$ units. \\
ZOOM & Adjust the magnification level of the displayed image or view. \\
TEXT & Type string $s$ into the focused input field (e.g., patient ID, study date, filter query). \\
SEGMENT & Create or edit a segmentation or region of interest (ROI) on the displayed medical image. \\
COMPLETE & Mark the workflow or task as finished. \\
\bottomrule
\end{tabular}%
}
\end{table}



\definecolor{head}{RGB}{242,246,255}

\begin{table}[t]
\caption{Dataset statistics of \textbf{\textit{CareFlow}}.}
\label{tab:careflow_stats}
\centering
\small
\setlength{\tabcolsep}{5pt}
\renewcommand{\arraystretch}{1.15}
\rowcolors{2}{white}{gray!3}

\resizebox{\columnwidth}{!}{%
\begin{tabular}{l|cc|cc|cc}
\rowcolor{head}
\toprule
\textbf{Split} & \textbf{\# Tasks} & \textbf{Avg.\ Steps} & \textbf{Min} & \textbf{Max} & \textbf{OOD} & \textbf{Actions} \\
\midrule
Train & 735 & 12.7 & 7  & 22 & -- & 6 \\
Test  & 315 & 12.9 & 9  & 24 & 50 & 6 \\
\midrule
\textit{Total} & 1050 & -- & -- & -- & 50 & 6 \\
\bottomrule
\end{tabular}%
}
\end{table}


\section{{\em CarePilot}}

Recent VLMs (GPT-4o, Gemini 2.5 Pro, Qwen VL 3) ground and perceive well in general settings but struggle on real healthcare software, with low task completion despite moderate step-wise accuracy. This limitation motivates \textbf{\textit{CarePilot}}, a framework that combines multimodal grounding, hierarchical reflection, and dual-memory reasoning to robustly automate complex clinical interfaces. The overall framework is shown in Figure \ref{fig:main_flow}.

\subsection{Task Definition}
Given a goal $g$ illustrated in natural language that requires a sequence of $T\!\in[t\_low,t\_high]$ steps in a healthcare software environment, the agent observes at each time $t$ the current screenshot $x_t$ and history $h_t$, and must choose a semantic action $a_t\!\in\!\mathcal{A}$ such that the overall sequence completes the task correctly. We formalize this as selecting actions that maximize execution success:
\begin{equation}
\hat{a}_{1:T}
=\ind\!\Big[\,V\!\big(g, x_{1:T}, a_{1:T}\big)=1\,\Big].
\end{equation}
where $V(\cdot)$ is a verifier that returns $1$ iff the workflow is successfully completed (i.e., all required artifacts and states are achieved).


\subsection{Tool Grounding}
To better parse healthcare visual interfaces inspired by \cite{wu2023visual}, we integrate four lightweight perception tools into the rollout and feed their outputs back to the MLLM for grounded next-action prediction: 
(1) \textbf{UI object detection (open-vocabulary)}: given a screenshot $\mathcal{I}_{\text{in}}$ and a text query $q$ (e.g., ``MPR,'' ``Export,'' ``Orders''), it returns bounding boxes $\mathcal{B}_{\text{out}}$ over matching widgets; 
(2) \textbf{Zoom/Crop}: from a region $\mathcal{R}$ on $\mathcal{I}_{\text{in}}$, it produces a magnified view $\mathcal{I}_{\text{focus}}$ to inspect small controls; 
(3) \textbf{OCR}: extracts token–box pairs $\mathcal{T}_{\text{out}}=\{(w_i,\mathbf{b}_i)\}$ for labels such as series names, patient fields, order IDs, and LIS codes, disambiguating visually similar elements; and 
(4) \textbf{Template/Icon matching}: given $\mathcal{I}_{\text{in}}$ and a template $\tau$ (e.g., measure/save/send-to-PACS), it returns matches $\mathcal{M}_{\text{out}}$ robust to themes, scaling, and locales. 
These four modules provide the best reliability benefit trade-off among tested toolsets.
The outputs of these four perception tools are aggregated into a unified representation denoted as $\phi_t$.
This tool-grounded feature vector $\phi_t$ serves as the perceptual grounding signal for subsequent modules, conditioning both memory updates and action prediction.

\subsection{Memory Utilization}
Long-horizon healthcare workflows require reasoning over both current and past contexts. 
Building on the perceptual grounding from the tool modules, CarePilot further introduces a dual-memory mechanism to reason over both current and past contexts in long-horizon workflows.

At each step $t$, the agent updates:
\begin{align}
\mathcal{M}^{S}_t &= f^{S}(x_{t-1}, a_{t-1}, r_{t-1}), \\
\mathcal{M}^{L}_t &= f^{L}\big(\mathcal{M}^{L}_{t-1}, \mathcal{M}^{S}_t, \phi_t\big),
\end{align}
where $\mathcal{M}^{S}_t$ denotes the \emph{short-term memory} summarizing the most recent context (previous screenshot, executed action, and critic feedback $r_{t-1}$), and $\mathcal{M}^{L}_t$ denotes the \emph{long-term memory}, a compact trajectory embedding updated using tool-grounding features $\phi_t$. The next action is conditioned on both memories:
\begin{equation}
a_t = \pi_\theta(g, x_t, \mathcal{M}^{S}_t, \mathcal{M}^{L}_t),
\end{equation}
where $\pi_\theta$ is the multimodal policy. This dual-memory mechanism stabilizes long-horizon reasoning, mitigates error accumulation, and preserves semantic consistency across workflows, consistent with prior hierarchical memory agents~\cite{sun2026h,iqbal2019actorattentioncritic}.
The resulting short- and long-term memories are then consumed by the Actor–Critic framework to condition future actions and guide reflection-based updates.

\subsection{Actor-Critic Framework}
Leveraging both the perceptual grounding from tools and the temporal context maintained in memory, the Actor–Critic framework forms the core decision module of CarePilot.
Both the Actor and Critic are instantiated from the same multimodal LLM (Qwen-VL~2.5-7B), differing only in their functional roles i.e., \emph{proposal} versus \emph{evaluation}, and their input conditioning. 

\definecolor{rowgray}{gray}{0.96}
\definecolor{headblue}{RGB}{240,245,255}
\definecolor{caregreen}{RGB}{230,250,235} 
\definecolor{closedgray}{RGB}{242,242,242}

\begin{table*}[t]
\centering
\small
\setlength{\tabcolsep}{5pt}
\renewcommand{\arraystretch}{1.0}
\rowcolors{2}{white}{white}

\caption{Results on \textbf{CareFlow}. Columns group \emph{Step-Wise Accuracy (SWA)} and \emph{Task Accuracy (TA)} for each software; the last group reports the overall \emph{Average}. Best results are \textbf{bold}, best among baselines are \underline{underlined}. \textcolor{teal!60!black}{Green-highlighted rows} denote our proposed method.}
\label{tab:careflow-horizontal}

\resizebox{\textwidth}{!}{%
\begin{tabular}{l *{10}{c}}
\rowcolor{headblue}
\toprule
\multirow{2}{*}{\textbf{Model}} &
\multicolumn{2}{c}{\textbf{Weasis}} &
\multicolumn{2}{c}{\textbf{3D Slicer}} &
\multicolumn{2}{c}{\textbf{Orthanc}} &
\multicolumn{2}{c}{\textbf{OpenEMR}} &
\multicolumn{2}{c}{\textbf{Average}} \\
\cmidrule(lr){2-3}\cmidrule(lr){4-5}\cmidrule(lr){6-7}\cmidrule(lr){8-9}\cmidrule(lr){10-11}
 & \textbf{SWA} & \textbf{TA}
 & \textbf{SWA} & \textbf{TA}
 & \textbf{SWA} & \textbf{TA}
 & \textbf{SWA} & \textbf{TA}
 & \textbf{SWA} & \textbf{TA} \\
\midrule
Qwen 2.5 VL 7B &  58.64 & 1.33  & 61.4 &  1.66 &  65.5 &  2.5 & 63.20  & 1.66 & 57.18 & 1.78 \\
Qwen 3 VL 8B & 65.20  &  1.33  &  68.38  &  3.33  &  67.71 & 5.0  & 66.31 &  5.0  & 66.90  & 3.66 \\
Qwen2.5 VL 32B & 79.95 & 5.13 & 68.00 & 1.90 & 48.62 & 2.42 & 41.60 & 0.32 & 60.72 & 2.43 \\
Llama 3.2 11B  & 86.03 & 10.26 & 76.47 & 4.60 & 69.58 & 13.58 & 62.56 & 11.47 & 75.65 & 9.50 \\
Llama 4 Scout  & 84.34 & 10.26 & 78.47 & 2.70 & \underline{88.56} & 23.90 & \underline{85.55} & 21.79 & \underline{85.65} & 13.50 \\
Llama 4 Maverick & 88.21 & 18.69 & 71.55 & 3.40 & 84.99 & 27.99 & 77.97 & 25.68 & 80.53 & 19.20 \\
Qwen3 VL 235B & 83.14 & 17.69 & 72.44 & 5.30 & 87.48 & 25.40 & 84.46 & 24.52 & 81.85 & 19.70 \\
Mistral3.2 VL 24B & 88.15 & 5.13 & 64.81 & 0.67 & 68.44 & 0.79 & 61.43 & 0.00 & 70.65 & 1.67 \\
Nemotron 12B VL & 86.98 & 12.82 & 73.95 & 5.13 & 73.56 & 14.46 & 66.55 & 12.36 & 77.93 & 10.71 \\
\rowcolor{closedgray}
GPT-4o & 85.30 & 20.0 & 77.5 & 27.37 & 88.5 & 26.67 & 85.1 & 27.50 & 83.13 & 25.40 \\
\rowcolor{closedgray}
GPT-5 & \underline{88.72} & \underline{31.25} & \underline{81.42} & \underline{37.9} & 86.92 & \underline{46.67} & 83.82 & \underline{31.25} & 85.22 & \underline{36.19} \\
\rowcolor{closedgray}
Gemini 2.5 Pro & 68.90 & 3.75 & 59.70 & 5.26 & 71.30 & 6.66 & 61.70 & 6.75 & 65.15 & 5.39 \\
\midrule
\rowcolor{caregreen}
\textbf{\textit{CarePilot (Qwen 2.5 VL-7B)}} &
\textcolor{ourteal}{90.38} & \textcolor{ourteal}{40.00} &
\textcolor{ourteal}{82.09} & \textcolor{ourteal}{54.75} &
\textcolor{ourteal}{93.80} & \textcolor{ourteal}{55.00} &
\textcolor{ourteal}{90.18} & \textcolor{ourteal}{56.70} &
\textcolor{ourteal}{88.05} & \textcolor{ourteal}{48.90} \\
\rowcolor{caregreen}
\textbf{\textit{CarePilot (Qwen 3 VL-8B)}} &
\textbf{\textcolor{ourteal}{92.50}} & \textbf{\textcolor{ourteal}{48.76}} &
\textbf{\textcolor{ourteal}{88.90}} & \textbf{\textcolor{ourteal}{54.80}} &
\textbf{\textcolor{ourteal}{91.80}} & \textbf{\textcolor{ourteal}{56.67}} &
\textbf{\textcolor{ourteal}{87.52}} & \textbf{\textcolor{ourteal}{46.25}} &
\textbf{\textcolor{ourteal}{90.18}} & \textbf{\textcolor{ourteal}{51.45}}\\
\bottomrule
\end{tabular}%
}
\end{table*}

At time $t$, the \textbf{Actor} observes $(x_t, g, \phi_t, \mathcal{M}^{S}_t, \mathcal{M}^{L}_t)$ and samples a semantic action:
\begin{equation}
a_t \sim \pi_{\theta}\!\big(a_t \mid x_t, g, \phi_t, \mathcal{M}^{S}_t, \mathcal{M}^{L}_t \big).
\end{equation}
The \textbf{Critic}, parameterized by $\phi$, evaluates the proposal via a value function:
\begin{equation}
Q_{\phi}(x_t, g, a_t, \mathcal{M}^{S}_t, \mathcal{M}^{L}_t) \rightarrow \hat{r}_t \in [0,1],
\end{equation}
where $\hat{r}_t$ estimates the action’s correctness. If $\hat{r}_t > \tau$, the Critic approves and updates both memories; otherwise, it issues structured feedback $\delta_t$ through hierarchical reflection.

\textbf{Hierarchical Reflection.}
If a prediction is incorrect ($\hat{r}_t \le \tau$), the Critic applies a three-level reflection: 
(i) the \emph{Action Reflector} compares consecutive states $(x_t, x_{t+1})$ to detect local grounding or perception errors; 
(ii) the \emph{Trajectory Reflector} inspects a short window $\{a_{t-k},\dots,a_t\}$ to diagnose stalled progress or violated preconditions; and 
(iii) the \emph{Global Reflector} evaluates the entire trajectory $\{a_1,\dots,a_t\}$ for goal consistency and decide if the task is completed yet or not. The action reflector is stored in the short-term memory, and the trajectory and global reflector gets stored in long-term memory.
The resulting feedback $\delta_t^{(S)}$ and $\delta_t^{(L)}$ update the corresponding memories:
\begin{align}
\mathcal{M}^{S}_{t+1} &= f^{S}(\mathcal{M}^{S}_t, a_t, \delta_t^{(S)}), \\
\mathcal{M}^{L}_{t+1} &= f^{L}(\mathcal{M}^{L}_t, \delta_t^{(L)}).
\end{align}
This hierarchical update promotes localized correction and long-term stability.
\subsection{Training Strategy}
After simulating actor-critic trajectories, we distill the Critic's feedback into the Actor following a reasoning distillation paradigm~\cite{sun2025chiron, qi2024cogcom}, eliminating the need for explicit multi-agent evaluation at inference time. The Actor is fine-tuned \textit{exclusively} on Critic-augmented \textit{successful} trajectories $\{(x_i, g_i, \phi_i, \mathcal{M}_i^{S}, \mathcal{M}_i^{L}, a_i^\star)\}_{i=1}^{N}$, where $a_i^\star$ denotes the Critic-verified and corrected next action. Each training sample also includes associated metadata, the updated memory state, and required tool-grounding information, which together form the Actor's full input context at step $t{+}1$. Because the Actor is trained only on verified successful trajectories, the feedback signal $r_{t-1}$ is always positive, and training follows a teacher-forced assumption in which all previous steps are assumed correct. This avoids any distribution shift during step-by-step autoregressive inference. The supervised fine-tuning loss is:
\begin{equation}
\mathcal{L}_{\text{SFT}}
= - \frac{1}{N} \sum_{i=1}^{N}
\log \pi_{\theta}\!\Big(a_i^\star \mid x_i, g_i, \phi_i, \mathcal{M}_i^{S}, \mathcal{M}_i^{L} \Big).
\end{equation}
At inference, only $\pi_\theta$ is retained: given the current GUI state, instruction, and memory context, the Actor directly predicts the next semantic action without any Critic involvement. In inference, the distilled Actor approximates the Critic’s reasoning, eliminating runtime overhead while preserving performance.
This design preserves the Critic's structured reasoning and memory usage within the Actor's parametric knowledge, enabling both faster inference and stronger performance compared to zero-shot and explicit actor-critic feedback loops.

\begin{figure*}[t]
    \centering
    \includegraphics[width=\linewidth]{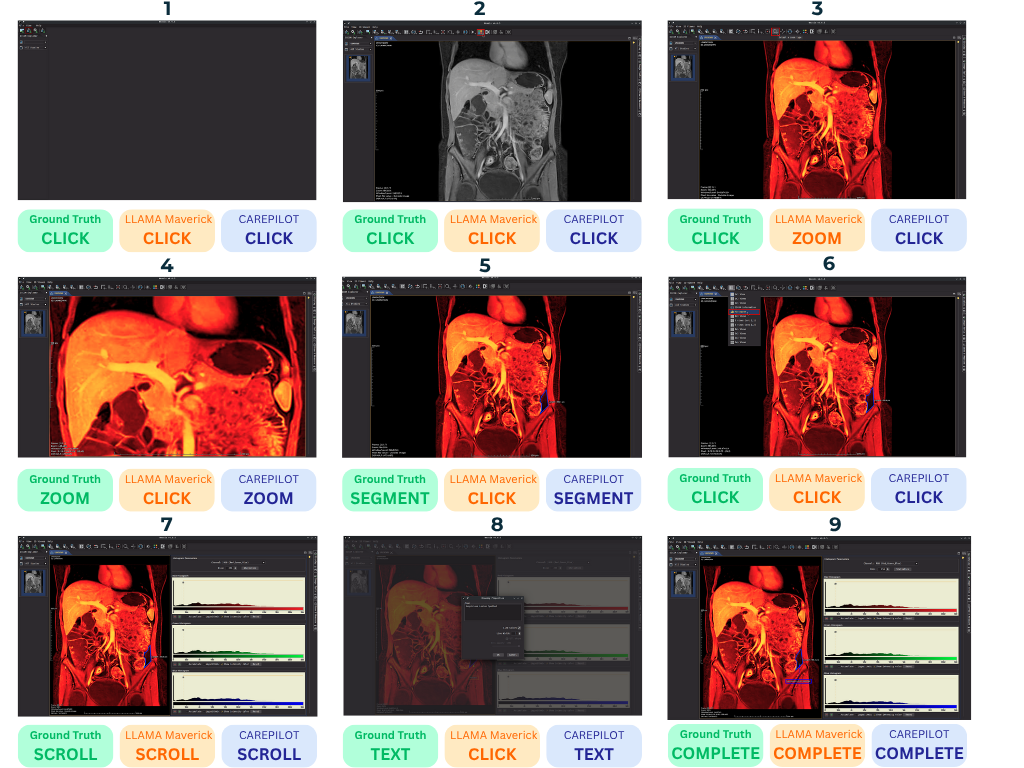}
    \caption{
    Qualitative visualization of \textbf{Llama-4 Maverick-17B} performing CarePilot’s radiology workflow tasks. 
    The traces highlight typical action–mode confusions such as issuing \texttt{ZOOM} instead of \texttt{CLICK} for tool selection and \texttt{CLICK} in place of \texttt{SEGMENT} or \texttt{TEXT} operations. 
    These incomplete branches illustrate the model’s inconsistent tool arming and gesture execution.}
    \label{fig:q1}
\end{figure*}

\section{Experimental Setup}
This section outlines our experimental setup, including implementation details (Sec.~5.1), dataset design (Sec.~5.2), baselines (Sec.~5.3), and evaluation metrics (Sec.~5.4).

\textbf{Implementation Details.} 
All experiments were conducted on NVIDIA A100~(40\,GB) GPUs and Google Colab~Pro+ environments, with each model trained for roughly 5--6~hours. 
Our framework was implemented using PyTorch~\cite{paszke2019pytorch}, Hugging~Face~Transformers~\cite{wolf2020transformers}, and Unsloth\footnote{https://github.com/unslothai/unsloth} for efficient fine-tuning, while baselines were accessed via the DeepInfra API\footnote{https://deepinfra.com/}. 
We used a cosine learning-rate schedule with warmup~100, learning rate \(2\times10^{-4}\), weight decay~0.01, and two training epochs. 
The sequence length was capped at~768 with gradient\_accumulation\_steps =32, yielding an effective batch size of \(32\times\) the number of GPUs. 
Mixed precision fp16=True, gradient checkpointing, and clipping max\_grad\_norm =1.0 were enabled. 
Checkpoints were saved per epoch, logs every~10~steps, and evaluation was disabled (evaluation\_strategy=no) to allow uninterrupted epoch-wise training. We fine tune Qwen vision–language backbones using lightweight LoRA adapters (rank 2, lora\_alpha 4, dropout 0.1) applied to attention and MLP projections, with base weights loaded in 4 bit precision for efficiency.



\textbf{Evaluation Metrics.} 
Performance was measured using two complementary metrics: \emph{Step-Wise Accuracy (SWA)} and \emph{Task Accuracy (TA)}. 
SWA measures the proportion of correct next-action predictions across all steps and tasks, counting a step as correct only when the predicted action exactly matches the annotated label among available options. 
TA measures the fraction of tasks for which the model predicted all \(n\) required actions correctly in order; a task is marked as successful only under this exact-match condition. 
Together, SWA reflects the reliability of the fine-grained step, while TA captures the success of the end-to-end workflow.

\textbf{Baselines.} To enable an extensive and fair evaluation, we compare both open source and closed source models. On the open source side, we include Qwen models\cite{qwen2.5,qwen3vl2025} (Qwen 2.5 VL 32B and Qwen 3 VL 235B), Llama models\cite{meta_llama4_blog2025} (Llama 4 Scout and Llama 4 Maverick), as well as Mistral 3.2 VL 24B\cite{rastogi2025magistral} and Nemotron 12B VL\cite{deshmukh2025nvidia}. Among closed source systems, we evaluate GPT 4o\cite{achiam2023gpt}, GPT 5\cite{openai_models2025}, and Gemini 2.5 Pro\cite{deepmind_gemini_models2025}. All baselines except the GPT ones are accessed via their respective DeepInfra API deployments and they are evaluated in zero shot setting.

\begin{table}[t]
\centering
\small
\setlength{\tabcolsep}{6pt}
\renewcommand{\arraystretch}{0.9}
\caption{Results on \textbf{OOD OpenHospital}. Best is \textbf{bold}, best among baselines are \underline{underlined}. 
Green rows denote our method (\textbf{\textit{CarePilot}}), gray rows indicate closed-source models, and white rows represent open-source baselines.}
\label{tab:careflow-average-only}

\resizebox{\columnwidth}{!}{%
\begin{tabular}{lcc}
\rowcolor{headblue}
\toprule
\textbf{Model} & \textbf{SWA} & \textbf{TA} \\
\midrule
Qwen2.5 VL 32B        & 71.74 & 12.72 \\
Llama 3.2 11B         & 70.76 & 16.36 \\
Llama 4 Scout         & 72.20 & 20.75 \\
Llama 4 Maverick      & 73.71 & 27.27 \\
Qwen3 VL 235B         & 75.18 & 25.46 \\
Mistral 3.2 VL 24B    & 69.63 & 1.82 \\
Nemotron 12B VL       & 72.90 & 18.18 \\
\rowcolor{rowgray}
Gemini 2.5 Pro        & 73.90 & 18.87 \\
\rowcolor{rowgray}
GPT-4o                & 74.63 & 25.48 \\
\rowcolor{rowgray}
GPT-5                 & \textbf{\underline{79.70}} & \underline{34.80} \\
\midrule
\rowcolor{caregreen}
\textbf{\textit{CarePilot (Qwen-VL 2.5-7B)}} & 77.93 & 36.40 \\
\rowcolor{caregreen}
\textbf{\textit{CarePilot (Qwen 3 VL-8B)}} & 79.27 & \textbf{38.18} \\
\bottomrule
\end{tabular}%
}
\end{table}

\section{Results and Findings}
We evaluate \textbf{\textit{CarePilot}} against strong \emph{open} and \emph{closed} source multimodal baselines across all healthcare domains, providing per–software class breakdowns to highlight strengths and weaknesses. The results, summarized in Table~\ref{tab:careflow-horizontal}, validate our hypothesis that tool grounding and dual (short- and long-term) memory significantly improve long-horizon performance.

\subsection{Research Questions}
\textbf{R1) How does \textbf{\textit{CarePilot}} perform compared to baselines?}  
\textbf{\textit{CarePilot}} consistently outperforms all open- and closed-source baselines across every healthcare domain. 
In \emph{task accuracy}~(TA), the Qwen~3~VL variant reaches 48.76\%, surpassing both its Qwen~2.5~VL counterpart (40.00\%) and the strongest baseline, GPT-5 (36.19\%). 
For \emph{step-wise accuracy}~(SWA), it achieves 92.50\%, exceeding Qwen~2.5~VL (90.38\%) and outperforming GPT-5 (85.22\%) by more than seven percentage points. 
These consistent gains highlight the impact of tool grounding, hierarchical feedback, and dual-memory design in improving long-horizon reasoning and interface generalization, as detailed in Table~\ref{tab:careflow-horizontal}.

\par
\definecolor{rowgray}{gray}{0.94}     
\definecolor{headblue}{RGB}{240,245,255}
\definecolor{caregreen}{RGB}{235,250,235} 

\textbf{R2) How is the performance of \textbf{\textit{CarePilot}} on Out of Distribution Healthcare Software benchmark?} On the Out of Distribution Open Hospital benchmark as shown in Table \ref{tab:careflow-average-only} , \textbf{\textit{CarePilot}} with Qwen 2.5 VL achieves an SWA of 77.93 and a task accuracy of 36.40, demonstrating superior robustness relative to strong open source and closed source models. Among open source baselines, Llama 4 Maverick is the strongest with a task accuracy of 27.27; among closed source models, the best reaches 34.80. Overall, the Qwen 3 VL version of \textbf{\textit{CarePilot}} is the strongest performer, underscoring its robustness on OOD tasks. 
\begin{table}[t]
\centering
\small
\setlength{\tabcolsep}{6pt}
\renewcommand{\arraystretch}{1.10}
\caption{Results to show the impact of Critic agent  in \textbf{\textit{CarePilot}} . WC represents without critic. (WC +TG) represents without critic agent but using tool grounding.}
\label{tab:careflow-critic-only}

\resizebox{0.8\columnwidth}{!}{%
\begin{tabular}{lcc}
\rowcolor{headblue}
\toprule
\textbf{Model} & \textbf{SWA} & \textbf{TA} \\
\midrule
CarePilot(WC)        & 65.37 &  3.75 \\
CarePilot(WC + TG)   &  72.98  & 12.5 \\
\midrule
\rowcolor{caregreen}
\textbf{\textit{CarePilot (Qwen-VL 2.5-7B)}} & \textbf{88.05} & \textbf{48.90} \\
\bottomrule
\end{tabular}%
}
\end{table}

\textbf{R3) How do open-source models compare to closed source models on \textbf{\textit{CareFlow}}?}  
We evaluate a broad set of open and closed-source multimodal agents, with overall results summarized in Table~\ref{tab:careflow-horizontal}. 
Among closed-source systems, GPT models achieve the highest average task accuracy (TA), reaching 36.19\%, while Gemini~2.5~Pro performs substantially lower. 
Within open-source models, the \texttt{Llama} family shows the strongest results, particularly \texttt{Llama-4~Scout}, which matches or surpasses several closed-source systems. 
Overall, closed-source GPT models maintain a performance lead among proprietary systems, whereas \texttt{Llama} variants dominate the open-source group, highlighting a narrowing gap between the two model classes on \textbf{\textit{CareFlow}}.\par

\textbf{R4)  How is the performance of \textbf{\textit{CarePilot}} without the Critic Agent? }
To understand the contribution of the Critic agent, we conduct experiment as shown in Table \ref{tab:careflow-critic-only} where we employ only the Actor agent under two configurations: one with access to tools and one without. In this setting, the task performance (TA) drops to 3.75 without tools and to 12.5 with tools, both of which are substantially lower than the performance of the full \textbf{\textit{CarePilot}} framework using the Qwen-2.5 VL 7B model. These findings indicate that the Critic’s role in facilitating hierarchical reflection is a key factor underpinning \textbf{\textit{CarePilot}}’s overall effectiveness.

\subsection{Ablation Studies}

\noindent\textbf{Impact of each component in \textbf{\textit{CarePilot}}.} 
We conduct extensive ablations in  Table~\ref{tab:history-overall-noindex} to quantify the contribution of each component in \textbf{\textit{CarePilot}}. Tool grounding (TG) emerges as the most critical: removing TG drops task accuracy to 9.37. This supports our conclusion that proper grounding supplies the context needed for accurate next-action prediction. We also assess the effects of removing Short-Term Memory (STM) and Long-Term Memory (LTM). In our evaluations,  LTM proves more consequential; removing it results in a larger performance decline than STM.

\noindent\textbf{Variance of results across different healthcare software.} Table \ref{tab:careflow-horizontal} shows the performance of models across four different healthcare softwares. Across all four software packages, 3D Slicer emerges as the most challenging setting for baselines (TA $\leq$ 5.3), likely due to longer, tool-dependent action chains; here, \textbf{\textit{CarePilot}} shows the largest gains, underscoring the central importance of tool grounding and memory. Although Orthanc and OpenEMR yield the highest baseline TAs (mid-teens to high-20s), \textbf{\textit{CarePilot}} still approximately doubles these results, and it maintains a similarly large margin on Weasis. \par

\noindent\textbf{Variation of Performance with Task Length.}  
Figure~\ref{fig:careflow_distribution} illustrates a clear performance decline as task length increases for both \textbf{\textit{CarePilot}} variants. 
For short workflows (<10 steps), accuracy remains high, above 64\% for both models. 
Between 10--15 steps, accuracy decreases notably, with a steeper drop for the 7B variant. 
Beyond 15 steps, the decline becomes pronounced: accuracy falls below 35\%, and for tasks exceeding 20 steps, both models converge around 27\%. 
Across all ranges, the Qwen~3~VL variant consistently outperforms Qwen~2.5~VL, demonstrating greater stability and resilience on longer, more complex sequences.

\begin{figure}[t]
    \centering
    \includegraphics[width=\linewidth]{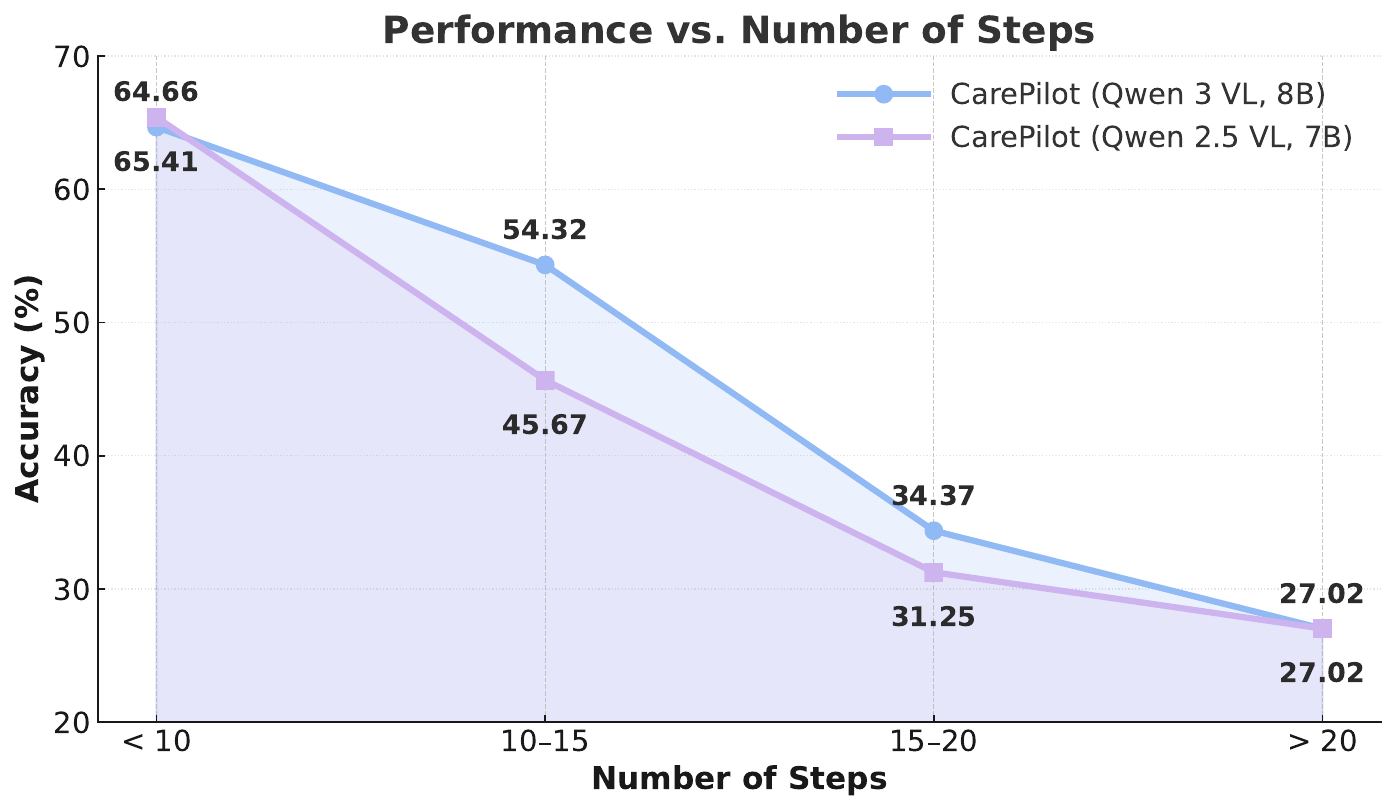}
    \caption{Variation of performance of  \textbf{\textit{CareFlow}} across different steps range.}
    \label{fig:careflow_distribution}
\end{figure}

\noindent\textbf{Qualitative Analysis.}  
To analyze baseline failures and \textbf{\textit{CarePilot}}’s gains, we examined representative task traces as shown in Figure-\ref{fig:q1}. 
Both \textbf{GPT-5} and \textbf{LLaMA-4~Maverick} succeed in initial setup and completion but often fail mid-sequence during \textit{mode switching}, confusing \texttt{CLICK} with \texttt{ZOOM} or \texttt{SCROLL}. 
\textbf{GPT-5} shows slightly better consistency, while \textbf{LLaMA-4~Maverick} misfires earlier, especially on segmentation and annotation steps. 
In contrast, \textbf{\textit{CarePilot}} maintains stable behavior by leveraging tool-grounding signals and memory-based feedback. These patterns highlight the importance of contextual reasoning and multi-step consistency for robust long-horizon execution.  Additional qualitative examples and visualizations are provided in the \textit{supplementary material}.\par

\newcommand{\cmark}{\textcolor{green!70!black}{\ding{51}}}   
\newcommand{\xmark}{\textcolor{red!80!black}{\ding{55}}}    
\definecolor{rowgray}{gray}{0.96}
\definecolor{caregreen}{RGB}{235,250,235} 

\begin{table}[t]
\centering
\footnotesize
\setlength{\tabcolsep}{8pt}
\renewcommand{\arraystretch}{0.7}
\caption{Ablation on contextual components using \textbf{\textit{CarePilot (Qwen 2.5 VL 7B)}}: Tool Grounding (TG), Long-Term Memory (LTM), and Short-Term Memory (STM). Reported are overall Step-Wise Accuracy (SWA) and Task Accuracy (TA). Best results are \textbf{bold}, and second-best are \underline{underlined}.}
\label{tab:history-overall-noindex}
\resizebox{\columnwidth}{!}{%
\begin{tabular}{ccc|cc}
\toprule
\multicolumn{3}{c|}{\textbf{Contextual Information}} &
\multicolumn{2}{c}{\textbf{Overall}} \\
\cmidrule(lr){1-3}\cmidrule(lr){4-5}
\textit{TG} & \textit{LTM} & \textit{STM} & \textbf{SWA} & \textbf{TA} \\
\midrule
\xmark & \cmark & \cmark & 73.20 & 9.37 \\
\rowcolor{white}
\cmark & \xmark & \cmark & \underline{82.10} &  23.67 \\
\cmark & \cmark & \xmark & 80.40 & \underline{30.42} \\
\rowcolor{caregreen}
\cmark & \cmark & \cmark & \textbf{88.05} & \textbf{48.90} \\
\bottomrule
\end{tabular}%
}
\end{table}

\noindent\textbf{Additional Supplementary Material.}  We provide additional details in the supplementary material, including qualitative analyses, the prompts used, further experiments on \textbf{\textit{CarePilot}} with the Qwen-3 8B-VL model, extended ablation studies, more information on data sources, inference-performance tradeoff analysis and the ethical considerations followed in this work.

\section{Conclusion}
We introduce \textbf{\textit{CarePilot}}, a novel multi-agent framework comprising an action and critic agent for automating long-horizon tasks in healthcare. It combines tool grounding with historical context through two complementary memory modules: a \emph{short-term memory} (STM) that stores the most recent step, outcome, and rationale, and a \emph{long-term memory} (LTM) that maintains trajectory-level context for reasoning across steps. We also present \textbf{\textit{CareFlow}}, the first benchmark dedicated to long-horizon healthcare software tasks, encompassing samples from multiple platforms across diverse clinical subdomains. Experiments demonstrate that \textbf{\textit{CarePilot}} achieves state-of-the-art results, outperforming strong open- and closed-source multimodal agents when equipped with accurate contextual grounding and memory-based reasoning.

\textbf{Limitation.} 
\textbf{\textit{CareFlow}} covers only five healthcare platforms and does not yet capture the full diversity of real-world clinical software. 
Moreover, \textbf{\textit{CarePilot}} predicts high-level semantic actions rather than exact GUI coordinates. 
Future work will expand platform coverage, add pixel-level grounding, and support longer, multilingual workflows.

\section{Acknowledgement}
Akash Ghosh would like to sincerely thank MBZUAI for providing the computational resources and infrastructure necessary to conduct the experiments. He also expresses his gratitude to Dr. Muhsin Muhsin, Academic Resident, Department of Community Medicine, IGIMS Patna, and Dr. Maleeka Zainab, Academic Resident, Department of Radiodiagnosis, PMCH, for their valuable guidance in the development of the dataset and validation of the CareFlow benchmark.
{
    \small
    \bibliographystyle{ieeenat_fullname}
    \bibliography{main}
}

\clearpage
\setcounter{page}{1}
\maketitlesupplementary
We include this supplementary document to further clarify certain sections of the main paper. It is organized as follows:\par
(i) Additional Experiments\par
(ii) Ethical Considerations\par
(iii) Qualitative Analysis\par
(iv) Information on Data Sources\par
(v) Inference Speed Analysis\par
(vi) Prompts.\par

\textbf{Additional Experiments}\par
We conducted additional experiments to assess the impact of different components of CarePilot on the Qwen 3 VL 8B model, as summarized in Table \ref{tab:Context Info Qwen3}. The results exhibit the same overall trend observed with the Qwen 2.5 VL variant: tool grounding contributes the most to performance, followed by long-term memory (LTM) and then short-term memory (STM).\par
To further understand the contribution of individual tools, we performed an ablation study, the results of which are presented in Table \ref{tab:tools-ablation}. We find that removing TM leads to the largest performance drop, highlighting its critical role, followed by OCR. In contrast, removing the zoom tool has the smallest impact, indicating that it is the least important component for our task.

\textbf{Ethical Considerations}\par
In this work, we collaborated closely with medical experts in the dataset-curation phase, ensuring that the images, annotations and clinical metadata were reviewed by domain-trained professionals. We commit to releasing our resulting datasets and trained models exclusively for non-commercial research use, aiming to advance scientific progress without commercial exploitation. All annotation and workflow tools employed (e.g., the DICOM viewer, image-computing platforms, open hospital/EMR systems) are publicly accessible and do not require proprietary licenses for use in this benchmark.\par

\begin{table}[t]
\centering
\footnotesize
\setlength{\tabcolsep}{8pt}
\renewcommand{\arraystretch}{0.7}
\caption{Ablation on contextual components using \textbf{\textit{CarePilot (Qwen 3 VL 8B)}}: Tool Grounding (TG), Long-Term Memory (LTM), and Short-Term Memory (STM). Reported are overall Step-Wise Accuracy (SWA) and Task Accuracy (TA). Best results are \textbf{bold}, and second-best are \underline{underlined}.}
\label{tab:Context Info Qwen3}
\resizebox{\columnwidth}{!}{%
\begin{tabular}{ccc|cc}
\toprule
\multicolumn{3}{c|}{\textbf{Contextual Information}} &
\multicolumn{2}{c}{\textbf{Overall}} \\
\cmidrule(lr){1-3}\cmidrule(lr){4-5}
\textit{TG} & \textit{LTM} & \textit{STM} & \textbf{SWA} & \textbf{TA} \\
\midrule
\xmark & \cmark & \cmark & 75.38 &  11.25  \\
\rowcolor{white}
\cmark & \xmark & \cmark & 84.22  &  26.25   \\ 
\cmark & \cmark & \xmark &   82.53  & 32.50 \\
\rowcolor{caregreen}
\cmark & \cmark & \cmark &  \textbf{90.18}  & \textbf{51.45}  \\
\bottomrule
\end{tabular}%
}
\end{table}

\begin{figure*}[t]
    \centering
    \includegraphics[width=0.88\linewidth]{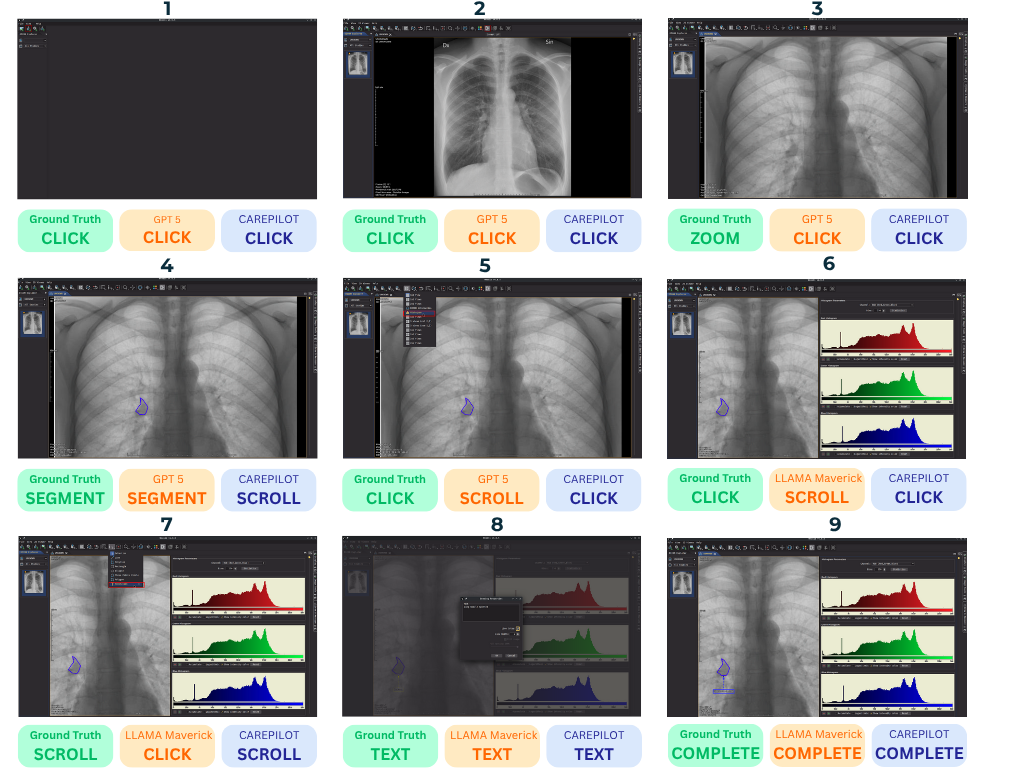}
    \caption{
    Qualitative visualization of \textbf{GPT-5} predictions for the same CarePilot tasks. 
    While GPT-5 improves on low-level action consistency, it still fails under domain-shifted UI states—issuing repeated \texttt{SCROLL} commands or premature free-text annotations before activating the correct tool. 
    These errors contrast with CarePilot’s complete, state-aware execution across both CT abdomen and chest workflows.}
    \label{fig:q2}
\end{figure*}

\begin{table}[t]
\centering
\small
\setlength{\tabcolsep}{8pt}
\renewcommand{\arraystretch}{0.7}
\caption{Ablation on tool components using \textbf{\textit{CarePilot (Qwen 2.5 VL 7B)}}: Object Detection (OD), Zoom (ZOOM), Optical Character Recognition (OCR), and Tool Memory (TM). Reported are overall Step-Wise Accuracy (SWA) and Task Accuracy (TA). Best results are \textbf{bold}, and second-best are \underline{underlined}.}
\label{tab:tools-ablation}
\resizebox{\columnwidth}{!}{%
\begin{tabular}{cccc|cc}
\toprule
\multicolumn{4}{c|}{\textbf{Tool Components}} &
\multicolumn{2}{c}{\textbf{Overall}} \\
\cmidrule(lr){1-4}\cmidrule(lr){5-6}
\textit{OD} & \textit{ZOOM} & \textit{OCR} & \textit{TM} &
\textbf{SWA} & \textbf{TA} \\
\midrule
\xmark & \cmark & \cmark & \cmark &  81.16 & 38.59  \\
\rowcolor{white}
\cmark & \xmark & \cmark & \cmark & 85.67  &  46.31 \\
\cmark & \cmark & \xmark & \cmark & 76.65  & 30.87 \\
\rowcolor{white}
\cmark & \cmark & \cmark & \xmark & 72.14  &  25.73 \\
\rowcolor{caregreen}
\cmark & \cmark & \cmark & \cmark & \textbf{88.05} & \textbf{48.90} \\
\bottomrule
\end{tabular}%
}
\end{table}
\textbf{Qualitative and Error Analysis}
\label{sec:qual}
We qualitatively compare the open-source \textbf{Llama-4 Maverick-17B} and the \textbf{GPT-5} baselines against our agent, \textbf{CarePilot}, on two routine radiology workflows: (i) CT abdomen with soft–tissue preset, zoom to liver, polygon ROI, statistics, and textual annotation; and (ii) CT chest with lung preset, zoom to right upper lobe, polygon ROI, statistics, and textual annotation. The visual traces in \autoref{fig:q1} (Llama) and \autoref{fig:q2} (GPT-5) show predicted actions overlaid on the UI sequence.

\paragraph{Where baselines fail (action–mode confusions).}
Both baselines repeatedly conflate \emph{tool selection} with \emph{tool execution}. In Task~1, Llama issues a \texttt{ZOOM} command when the ground truth requires a final \texttt{CLICK} to select the zoom tool (Step~5), and then attempts a \texttt{CLICK} where the correct action is a drag-based \texttt{ZOOM} gesture (Step~6). It similarly emits a \texttt{CLICK} instead of \texttt{SEGMENT} for polygon drawing (Step~8), and \texttt{CLICK} instead of \texttt{TEXT} for annotation (Step~11). GPT-5 exhibits the same pattern: \texttt{CLICK} in place of \texttt{ZOOM} (Task~1, Step~6; Task~2, Step~5) and \texttt{SCROLL} or \texttt{CLICK} in place of \texttt{SEGMENT} or opening \texttt{STATS} (Task~1, Steps~8–9; Task~2, Steps~7–9). These errors manifest as short, incorrect branches in the traces in \autoref{fig:q1}–\autoref{fig:q2} (missed zoom gesture, incomplete polygon, stats panel not invoked).

\paragraph{State–awareness and panel navigation errors.}
A second class of failures arises from poor \emph{UI state tracking}. After opening annotation palettes, both baselines occasionally behave as if they were still in navigation mode, issuing \texttt{SCROLL} where a targeted \texttt{CLICK} on the statistics tool is required (GPT-5: Task~1 Step~9; Task~2 Steps~8–9). This suggests the policies are not verifying latent UI mode (navigation vs.\ annotation) before acting.

\paragraph{Premature free-text emission.}
In Task 2, GPT-5 produces free text (e.g., \emph{``Pulmonary Nodule- 8\,mm''}) \textbf{before} the text tool is properly armed (Step~10), and then repeats the text once the tool is finally active (Step~11). This behavior indicates weak coupling between language generation and GUI affordances.

\paragraph{CarePilot: consistent step completion.}
In contrast, \textbf{CarePilot} completes \emph{all} steps across both tasks (100\% step completion in the showcased cases). We attribute this to three design choices:
\begin{itemize}
    \item \textbf{Action–mode verification:} before executing a gesture, CarePilot explicitly checks the active tool state; if mismatched, it first issues the required selection \texttt{CLICK} and only then performs the gesture (\texttt{ZOOM} or \texttt{SEGMENT}).
    \item \textbf{UI-aware planning:} a short-horizon controller constrains admissible next actions by the visible widgets (e.g., stats icon present $\Rightarrow$ prioritize \texttt{CLICK} over \texttt{SCROLL}).
    \item \textbf{Grounded annotation:} text emission is gated on the text-tool cursor state, avoiding premature free-text.
\end{itemize}

\paragraph{Clinical relevance.}
These qualitative differences have a practical impact: missing a zoom gesture or failing to close a polygon yields incorrect ROI statistics, while premature annotation risks inconsistent reporting. CarePilot’s reliable tool arm, gesture execution, and panel navigation produce correct ROIs and measurements on first attempt, reducing operator friction and potential measurement variance.

\paragraph{Takeaways.}
As illustrated in \autoref{fig:q1} and \autoref{fig:q2}, the dominant baseline errors are (i) action–mode confusions, (ii) missing state checks for palette/panel transitions, and (iii) ungrounded text actions. CarePilot eliminates these classes through explicit state verification and affordance-aware planning, resulting in consistent end-to-end task completion.

\textbf{Information on Data Sources}\par

\textit{Orthanc:} Orthanc is a lightweight open-source DICOM server initially released in 2012, developed at the Université catholique de Louvain. It is designed to simplify the management of medical image workflows by providing a RESTful API on top of DICOM storage.  Use cases include setting up a mini-PACS, automating DICOM image transfers (C-STORE, C-FIND) and enabling research or departmental image archival. To use it, one installs Orthanc (often via Docker), configures its DICOM AE titles, sets up storage/back-end plugins, and interacts through its web UI or REST endpoints for querying and retrieving DICOM files.\par
\textit{3D Slicer:} 3D Slicer is a free, open-source platform for medical image computing, visualization, segmentation and registration, first developed from a 1998 master’s thesis project.  It is widely used in research for image-guided therapy, volume rendering, and custom algorithm prototyping.  To use it, one downloads the appropriate build (Windows, Linux, macOS), loads DICOM or other image volumes, uses modules/plugins for e.g., segmentation or registration, and exports results or integrates custom code via Python or C++.\par

\textit{OpenEMR:} OpenEMR is a widely-used open-source electronic health record (EHR) and practice management system, publicly launched under GPL in 2002. It supports scheduling, billing, and clinical records and has been certified for Meaningful Use in the US; deployed globally across clinics and hospitals. Use cases include managing patient demographics, tracking visits and tests, and generating invoices. To use it, one installs on a LAMP stack (Linux/Apache/MySQL/PHP), configures user roles and modules, customises forms/fields, and then staff access it via web browser.\par
\textit{OpenHospital:} OpenHospital is a free open-source hospital information system (HIS) designed especially for centres in low-resource settings, first deployed around 2006.  It supports patient registration, admissions, lab management, pharmaceuticals, and basic statistics. Use cases include managing day-to-day hospital workflows and laboratory operations in small to medium institutions. To use it, one installs the Java-based application (desktop or client/server), sets up the database, configures units and users, and uses its UI for managing patients, labs, and reports.\par
\begin{table}[t]
\centering
\footnotesize
\begin{minipage}{0.58\linewidth}
\centering
{\setlength{\tabcolsep}{4pt}\renewcommand{\arraystretch}{0.95}
\begin{tabular}{lcc}
\toprule
\textbf{Method} & \textbf{Avg. Time / Task} & \textbf{TWA} \\
\midrule
Qwen 2.5 VL (Zero Shot) & $\sim$20 s & 8.5 \\
 Actor-Critic, tools & $\sim$150 s & 42.5 \\
CarePilot (distilled) & \textbf{$\sim$90 s} & 48.9 \\
\bottomrule
\end{tabular}}
\end{minipage}

\caption{Empirical average inference time per long-horizon task on CareFlow(315 tasks).}
\label{tab:empirical_inference_time}

\end{table}

\textbf{Inference Speed Analysis}\par
To understand the impact of adaptation and the removal of the critic agent during inference, we analyze the trade-off between inference cost and performance under three settings: (i) zero-shot inference, (ii) vanilla actor–critic loops, and (iii) the proposed \textbf{\textit{CarePilot}} framework, as shown in Table X. Our analysis as shown in Table \ref{tab:empirical_inference_time} indicates that \textbf{\textit{CarePilot}} achieves significantly stronger performance compared to the vanilla actor–critic framework with iterative loops, while maintaining a more efficient inference procedure.

\textbf{Prompts}\par

\begin{tcolorbox}[
  breakable,
  enhanced,
  width=\linewidth,   
  colback=blue!5!white,
  colframe=blue!75!black,
  title=Prompt for Actor Execution Agent,
]

\textit{\textbf{You are the TARGET EXECUTION AGENT controlling a mobile/GUI environment step by step.}}

You can use VISUAL GROUNDING TOOLS to better understand the screen:
\begin{itemize}
  \item \textbf{object\_detection}: Find UI elements like buttons, icons, text fields
  \item \textbf{visual\_grounding}: Find specific elements by text query (e.g., ``Load Data button'')
  \item \textbf{depth\_estimation}: Understand UI hierarchy and layering
  \item \textbf{edge\_detection}: Identify UI boundaries
  \item \textbf{zoom\_tool}: Zoom into specific regions for detailed inspection
\end{itemize}

If you're uncertain about UI element locations or need precise coordinates, you \textbf{MUST} request tools in \texttt{"reasoning.tool\_calls"}. Tools will be executed and results provided back to you.

You \textbf{MUST} respond with \textbf{EXACTLY ONE} valid JSON list, no extra text, no explanations, no markdown fences.

\medskip

\noindent\textbf{Format:}
\begin{verbatim}
[
  "Step {step_num}" : {
    "grounding": {
      "current_screen_state": "...",
      "key_ui_elements": ["...","..."],
      "relevant_affordances": ["..."]
    },
    "short_term_memory": {
      "last_action": "...",
      "last_observation": "...",
      "last_lesson": "..."
    },
    "long_term_memory": {
      "overall_progress": "...",
      "completed_subtasks": ["..."],
      "remaining_subtasks": ["..."],
      "known_pitfalls": ["..."]
    },
    "reasoning": {
    "tool_calls": [
    {"tool": "visual_grounding",
        "args": {
        "query": "Load Data button", 
        "image_id": 0}},
        {"tool": "object_detection", 
        "args": {"objects": ["button", 
        "icon"]}}
      ],
      "why_next_action_is_correct
      _and_safe": "...",
      "why_it_aligns_with_user_goal"
      : "...",
      "why_alternatives_are
      _wrong_or_risky": "..."
    },
    "tool_results": {
      "visual_grounding": {...},
      "object_detection": {...}
    },
    "image_info": {
      "step_num": {step_num},
      "has_image": true,
      "image_data_uri": "{image
      _data_uri}"
    },
    "predicted_next_action": {
      "tool_call": "ONE_OF_
      AVAILABLE_TOOLS",
      "target": "UI element/selector
      / coords to operate on",
      "target_id": "element_id_
      from_ui_tree",  # MUST use an 
      ID from ui_tree if available
      "arguments": {
        "text_to_type": "...",
        "coords": [x, y],
        "extra": "..."
      }
    }
  }
]
\end{verbatim}

\medskip

\noindent\textbf{Constraints:}
\begin{enumerate}
  \item \textbf{"grounding"}: ONLY describe what is visible RIGHT NOW on THE CURRENT SCREEN. Do not invent elements.
  \item \textbf{"short\_term\_memory"}: ONLY summarize what happened in the immediately previous attempt (the last step), including \texttt{last\_action}, what we observed, and the immediate lesson. If step \texttt{\{step\_num\}} is the first step, use the given values (like ``NONE'').
  \item \textbf{"long\_term\_memory"}: Summarize cumulative progress in this task so far:
    \begin{itemize}
      \item what subgoals are already done,
      \item what remains,
      \item known pitfalls (e.g. loops or dead ends we discovered),
      \item \texttt{overall\_progress} so far.
    \end{itemize}
    If this is the first step, keep them minimal/empty.
  \item \textbf{"reasoning"}:
    \begin{itemize}
      \item \textbf{"tool\_calls"} (REQUIRED): You MUST call \texttt{visual\_grounding} or \texttt{object\_detection} if you need to interact with UI elements.
      \item Explain why the chosen next action is safe, aligned with \texttt{USER\_GOAL}, and better than other visible actions.
    \end{itemize}
  \item \textbf{"tool\_results"} (OPTIONAL): Will be populated by system after tool execution.
  \item \textbf{"predicted\_next\_action"}:
    \begin{itemize}
      \item \texttt{tool\_call} MUST be one of \texttt{AVAILABLE\_TOOLS}.
      \item You MUST provide all arguments that tool needs (coords, text, selector, etc).
      \item If \texttt{ui\_tree} is available, you MUST use \texttt{target\_id} from \texttt{ui\_tree} instead of free-text target.
      \item If \texttt{ui\_tree} is not available, you MUST request \texttt{visual\_grounding} in \texttt{tool\_calls}.
    \end{itemize}
  \item DO NOT add any keys not listed.
  \item DO NOT output anything except the JSON list described above.
  \item NEVER guess coordinates - if you need coords, you MUST use \texttt{visual\_grounding} tool first.
\end{enumerate}

\medskip

\noindent\textbf{User Prompt:}
\begin{verbatim}
USER_GOAL: {user_goal}

AVAILABLE_TOOLS: {formatted_tools}

SHORT_TERM_MEMORY 
(what happened in the previous step):
{json.dumps(stm, indent=2)}

LONG_TERM_MEMORY 
(cumulative knowledge and progress):
{json.dumps(ltm, indent=2)}
{tool_context}
{tool_effectiveness_hints}

Step {step_num}:
Based on the USER_GOAL, 
AVAILABLE_TOOLS, SHORT_TERM_MEMORY, 
LONG_TERM_MEMORY,and the current
screen, determine the next action.
\end{verbatim}

\end{tcolorbox}

\begin{tcolorbox}[
  colback=red!5!white,
  colframe=red!75!black,
  title=Prompt for Critic / Hierarchical Reflector Agent,
  enhanced,
  breakable
]

\textit{\textbf{You are the CRITIC / HIERARCHICAL REFLECTOR Agent}}

You must:
\begin{enumerate}
  \item Judge whether the Target Agent's
        \texttt{predicted\_next\_action} was
        correct in practice.
  \item Evaluate tool usage:
        Were appropriate tools called?
        Were results correctly interpreted?
  \item Produce step-level reflection
        (\texttt{reflection.action}).
  \item Produce trajectory-level reflection
        (\texttt{reflection.trajectory}).
  \item Produce global task-level reflection
        (\texttt{reflection.global}).
  \item Indicate \texttt{action\_correct}
        as true/false.
  \item If false, explain
        \texttt{why\_if\_wrong} and give
        \texttt{hint\_if\_wrong}.
  \item Provide \texttt{tool\_evaluation}
        with \texttt{tools\_used},
        \texttt{tool\_success}, and
        \texttt{tool\_lessons}.
\end{enumerate}

\medskip
\noindent\textbf{CRITICAL FORMAT REQUIREMENTS:}
\begin{itemize}
  \item \texttt{reflection.trajectory.
  completed\_subtasks}
        MUST be a JSON array of strings,
        e.g.,
        \texttt{["Load MRI data",
        "Navigate to module"]}.
  \item \texttt{reflection.trajectory.
  remaining\_subtasks}
        MUST be a JSON array of strings,
        e.g.,
        \texttt{["Create segmentation",
        "Export results"]}.
  \item NEVER use strings or text
        descriptions in place of arrays.
  \item Each subtask should be a short,
        specific task description
        (1--5 words).
  \item Extract subtasks from the
        \texttt{USER\_GOAL} based on actual
        progress made so far.
  \item \texttt{completed\_subtasks}:
        List what has been accomplished
        based on steps taken.
  \item \texttt{remaining\_subtasks}:
        List what still needs to be done
        based on remaining steps.
\end{itemize}

\medskip
\noindent\textbf{Subtask style:}
\begin{itemize}
  \item Each subtask should be short
        (1--5 words), specific, and
        actionable.
  \item Do \emph{not} use long descriptive
        sentences; use short, specific
        task names.
  \item Based on the trajectory, identify
        which parts of the
        \texttt{USER\_GOAL} have been
        completed.
\end{itemize}

\medskip
\noindent\textbf{CRITICAL COMPLETE ACTION RULES:}
\begin{itemize}
  \item The \texttt{"COMPLETE"} action
        can ONLY be used in the LAST STEP.
  \item If the agent used
        \texttt{"COMPLETE"} before the
        last step, mark
        \texttt{action\_correct} as FALSE.
  \item The \texttt{"COMPLETE"} action
        should ONLY appear when all
        required subtasks are truly
        finished.
  \item Never mark a task as
        \texttt{"complete"} in
        \texttt{reflection.global.status}
        unless it is actually the final
        step and all objectives are
        achieved.
  \item If there are
        \texttt{remaining\_subtasks} or
        \texttt{missing\_steps}, the
        status MUST be
        \texttt{"incomplete"}.
\end{itemize}

\medskip
\noindent You MUST return EXACTLY ONE JSON
object with the required keys including
\texttt{tool\_evaluation}. Do NOT include
any text outside that JSON.

\medskip
\noindent Evaluate both the ACTION and TOOL
USAGE:
\begin{itemize}
  \item Did the agent use appropriate
        tools?
  \item Were tool results correctly
        interpreted?
  \item Could better tools have been
        chosen?
  \item Did the tools help or hinder the
        action?
\end{itemize}

\medskip
\noindent\textbf{User Prompt Template:}

\begin{verbatim}
USER_GOAL:
{user_goal}

TARGET_AGENT_STEP_OUTPUT (Step 
{step_num}):
{target_output_json}

{tool_context}{memory_context}

GROUND_TRUTH_AFTER_ACTION
(what actually happened 
after executing
predicted_next_action):
{ground_truth_after_action}

FULL_TRAJECTORY_SO_FAR
(chronological summary of all steps
so far,
including this one):
{full_trajectory_so_far}

Now respond with the single JSON 
object
exactly in the required format,
including tool_evaluation.
\end{verbatim}

\end{tcolorbox}


\begin{tcolorbox}[
  colback=green!5!white,
  colframe=green!75!black,
  title=Prompt for  running Baselines in zero shot,
  enhanced,
  breakable
]

\textit{\textbf{Given a screenshot and an
instruction, provide the correct action.}}

\medskip
\noindent\textbf{Available Actions:}
\texttt{\{available\_action\_description\}}

\medskip
\noindent\textbf{IMPORTANT RULES — Choose the
correct action based on what you see:}

\begin{enumerate}
  \item \textbf{COMPLETE:}
        Only allowed on the final step
        (step \texttt{\{total\_steps\}}).
        Never use before the last step.

  \item \textbf{CLICK:}
        Use when interacting with UI
        elements:
        \begin{itemize}
          \item Buttons, menus, tool icons
          \item Highlights appear on UI
                tools/icons, \emph{not} on
                the medical scan
          \item Used for navigating,
                selecting tools, opening
                menus
        \end{itemize}

  \item \textbf{SEGMENT:}
        Use when annotations appear
        \emph{on the medical scan}:
        \begin{itemize}
          \item Points, fiducials,
                masks, measurements
          \item Lines, shapes, bounding
                boxes on MRI/CT images
          \item If annotation is on the
                scan itself →
                \textbf{SEGMENT}, not
                CLICK
        \end{itemize}

  \item \textbf{ZOOM:}
        Use when magnification of the
        medical scan changes:
        \begin{itemize}
          \item Scan becomes larger or
                smaller
          \item Zoom percentage changes
          \item No new annotations added
        \end{itemize}

  \item \textbf{TEXT:}
        Use when typing into input
        fields:
        \begin{itemize}
          \item Cursor active in a text
                box
          \item Text being entered
        \end{itemize}

  \item \textbf{SCROLL:}
        Use when vertical scrolling
        occurs:
        \begin{itemize}
          \item Content moves up/down
          \item Scrollbar changes
          \item New content becomes
                visible
        \end{itemize}

\end{enumerate}

\medskip
\noindent\textbf{Current Step:}
\texttt{\{current\_step + 1\}} of
\texttt{\{total\_steps\}}

\medskip
\noindent\textbf{Grounding Context:}
\texttt{\{grounding\_context\}}

\medskip
\noindent\textbf{Historical Actions:}
\texttt{\{history\}}

\medskip
\noindent\textbf{Instruction:}  
Based on the screenshot and the available
actions, provide the next step directly.

\textbf{Output ONLY the action type:}  
CLICK, SEGMENT, TEXT, SCROLL, or COMPLETE.

No coordinates.  
No explanations.  
\textbf{COMPLETE only on the last step.}

\end{tcolorbox}

\end{document}